%% file: main.tex
\documentclass{article}
\usepackage[accepted]{icml2019arxiv}

\usepackage{microtype}
\usepackage{graphicx}
\usepackage{booktabs} 

\usepackage[hyphens]{url}
\usepackage{color-edits}
\usepackage{amssymb}
\usepackage{lineno}
\usepackage{float}
\usepackage{cleveref}
\usepackage{authblk}
\usepackage{xspace}
\usepackage{amsmath,amsfonts}
\usepackage{xspace}
\usepackage{dsfont}
\usepackage{multirow}
\addauthor{jm}{red}
\newtheorem{remark}{Remark}

\addauthor{bw}{blue}
\addauthor{jh}{green}

\newcommand{\ls}{\texttt{LS}\xspace}
\newcommand{\ds}{\texttt{DS}\xspace}

\usepackage{hyperref}

\icmltitlerunning{Predictive Inequity in Pedestrian Detection}

\begin{document}

\twocolumn[

\icmltitle{Predictive Inequity in Object Detection}

\thispagestyle{plain}
\pagestyle{plain}


\icmlsetsymbol{equal}{*}

\begin{icmlauthorlist}

\icmlauthor{Benjamin Wilson}{gt}
\icmlauthor{Judy Hoffman}{gt}
\icmlauthor{Jamie Morgenstern}{gt}
\end{icmlauthorlist}

\icmlaffiliation{gt}{Georgia Tech}

\icmlcorrespondingauthor{Benjamin Wilson}{benjaminrwilson@gatech.edu}

\icmlkeywords{Machine Learning, ICML}

\vskip 0.3in
]



\printAffiliationsAndNotice{}  


\begin{abstract}
In this work, we investigate whether state-of-the-art object detection systems have \emph{equitable predictive performance} on pedestrians with different skin tones. This work is motivated by many recent examples of ML and vision systems displaying higher error rates for certain demographic groups than others. We annotate an existing large scale dataset which contains pedestrians, BDD100K, with Fitzpatrick skin tones in ranges [1-3] or [4-6]. We then provide an in depth comparative analysis of performance between these two skin tone groupings, finding that neither time of day nor occlusion explain this behavior, suggesting this disparity is not merely the result of pedestrians in the 4-6 range appearing in  more difficult scenes for detection. We investigate to what extent time of day, occlusion, and reweighting the supervised loss during training affect this predictive bias.
\end{abstract}

\input{intro.tex}
\input{prelim}

\input{evaluation.tex}

\input{conclusion.tex}
\bibliography{citations}
\bibliographystyle{icml2019}

\newpage
\newpage
\pagebreak

\input{appendix}

\end{document}

%% file: intro.tex
\section{Introduction}

The methods and models developed by the machine learning community have begun to find homes 
throughout our daily lives: they shape what news stories and advertisements we see online, the engineering of new products sold in stores, the content of our emails, and, increasingly, the allocation of resources and surveillance. Both private and governmental organizations have increasingly begun to use such statistical methods. Examples of the latter include predictive policing, recidivism prediction, and the allocation of social welfare resources~\citep{predpol,recidivism,welfare,child}. A particularly driving application domain for ML in the private sector is the design of autonomous vehicles. Autonomous vehicles may greatly reduce transit costs of goods and reduce individuals' reliance on owning personal vehicles.

Recognizing key objects such as pedestrians and road signs  plays a key role in these systems, helping determine when a car must brake or swerve to avoid fatalities. The few autonomous vehicle systems already on the road have shown an inability to entirely mitigate risks of pedestrian fatalities~\citep{uber}. A natural question to ask is \emph{which} pedestrians these systems detect with lower fidelity, and why they display this behavior. In this paper, we study the performance of several models used in state-of-the-art~\citep{he2017mask} object detection, and show \emph{uniformly poorer performance of these systems when detecting pedestrians with Fitzpatrick skin types between 4 and 6.} This behavior suggests that future errors made by autonomous vehicles may not be evenly distributed across different
demographic groups.

We then investigate \emph{why} standard object detection 
might have higher predictive accuracy for pedestrians lower on the Fitzpatrick scale. The training set has roughly 3.5 times as many examples of  lower-Fitzpatrick scored pedestrians compared to higher scored pedestrians, which suggests several different sources of predictive disparity between the two groups might be at work. First, one would expect to have lower generalization error on the larger subset of data. Second, many standard loss functions will prioritize accuracy on the larger subset of the data. 

These two behaviors, and others, are often conflated and described by
both industry and researchers as ``biased data", a 
shorthand for a milieu of different issues of sampling, measurement, and weighting of different design goals. Sampling issues arise when a dataset does a poor job representing the larger population (e.g., a dataset with mostly men, or no examples of women who successfully repaid mortgages). Issues of measurement arise when the features collected are insufficient to accurately measure and predict the intended outcome variable (such as banking records having insufficient information to predict creditworthiness in ``unbanked" communities, where participation in lending circles~\citep{mccarthur} and other less centralized systems better predict loan repayment).  Designing an objective function (and constraints) for training an ML system tacitly weights different model behaviors over others, such as the weighting of false positives versus false negatives. Relatively little work in the fair ML community has explicitly teased apart these sources for particular examples of inequitable predictive behavior. {\bf We explicitly aim to measure three possible sources of predictive imbalance: whether time of day, occlusion, or if the loss function prioritizing accuracy on the larger population heavily impacts this behavior. }

\section{Related Work}

Predictive disparities of ML systems have recently been given much attention. These examples appear in numerous domains, a few of which we mention here. Advertising systems show ads based upon numerous demographic features; as a result, a number of findings have shown certain gender or racial groups receive certain ads at much higher rates than others~\citep{zhao2017men,datta2018discrimination,sweeney}. Certain predictive policing systems has been shown to have differing predictive performance based on race~\citep{propublica,selbst2017disparate,lum}. 

Most closely related to our current work are examples of vision-based systems with differing predictive qualities for women or people of color. Facial recognition systems (and other systems which use facial images) have garnered the lion's share of the press in this space. 

Early warnings that facial recognition might have higher accuracy on white men showed that this problem might be somewhat mitigated by training systems separately for different demographic groups~\citep{klare2012face}. Nevertheless, recent, state-of-the art systems 
designed by many major tech conglomerates have continued to face scrutiny for the behavior of their facial recognition systems. Commercial gender prediction software has been shown to have much worse accuracy on women with Fitzpatrick skin types 4-6 compared to other groups~\citep{gender_shades_2018}; this work inspired our use of the Fitzpatrick skin scale to categorize pedestrians. The ACLU found that Amazon's facial recognition system incorrectly matched a number of darker-skinned members of congress to mugshots from arrests across the country~\citep{snow_2018}.

Our work instigates the measurement of predictive imbalance for a different set of ML-guided vision systems, namely that of driving-centric object detection. This work is particularly timely as several locations have recently allowed autonomous vehicles to operate on public roads, several casualties have resulted. We do not focus on the ethical dilemmas algorithms might ultimately face\footnote{Should the car crash into the person on the left or right, if those are the only options? Several papers have investigated such questions, e.g.~\citet{trolly}. }, but instead on the simpler question of whether several simple building blocks used for research-grade pedestrian detection have similar ability to detect pedestrians with different skin tones. 

We study the problem of pedestrian detection in road scenes from the perspective of an autonomous vehicle. Many datasets have been introduced in the computer vision community for developing methods for recognizing pedestrians at a variety of distances~\citep{kitti,eth,Dollar2012PAMI} and for recognizing all objects in road scenes relevant to the autonomous driving task~\citep{cityscapes,yu2018bdd100k}. State-of-the-art region-proposal based methods for general object detection such as Faster R-CNN~\citep{ren2015faster} or Mask R-CNN~\citep{he2017mask} often form the backbone for the best performing pedestrian detection models for unoccluded pedestrians close to the vehicle for which more pixels are available for recognition. While the same region proposal based method can be effective for proposing bounding boxes for small (distant) pedestrians~\citep{he_pedestriandetection_2016}, often unique representations are needed for simultaneous detection of small (distant), large (nearby), and occluded pedestrians. Recent works have addressed this is by incorporating multiple representations into their model, one for each pedestrian scale~\citep{li_scaleaware_2018}, or for different body parts to handle occlusion~\citep{Zhang_2018_CVPR}.
In this work we compare detection performance of nearby and largely unoccluded pedestrians for which skin-tone is readily identifiable and therefore focus our analysis on the core technology that persists between object and pedestrian detection systems. 

%% file: prelim.tex
\section{Preliminaries}\label{sec:prelim}
We begin with an overview of the main concepts used in this work: the problem of pedestrian detection; the classification of people into groups based on skin tone and other characteristics known as Fitzpatrick skin typing; and predictive disparity, or the difference in predictive performance of a learning system on two different groups of datapoints.

\textbf{Pedestrian Detection} Quickly identifying pedestrians has been a long-standing challenge in the Computer Vision community. From security systems to autonomous cars, Pedestrian Detection remains an important and crucial aspect of many Computer Vision models. Common challenges of Pedestrian Detection include: occlusion by other objects or people, changes in clothing, and diverse lighting conditions.

\textbf{Fitzpatrick Skin Typing}
The Fitzpatrick skin type scale~\citep{fitzpatrick}, introduced to predict a person's predisposition to burning when exposed to UV light, measures a number of physical attributes of a person including skin, eye, and hair color, as well as a person's likelihood to freckle, burn, or tan. As a general rule, categories 1-3 correspond to lighter skin tones than 4-6.  This categorization aims to design a culture-independent measurement of skin's predisposition to burn, which correlates with the pigmentation of skin. 

\textbf{Predictive Inequity}
 Assuming a fixed partition of the person class (for example, into classes \ls and \ds based on the Fitzpatrick skin scale), we define the \emph{predictive inequity} of a model $f$ for a particular loss function $\ell$ to be the difference in loss the model $f$ incurs on members of \ls over \ds:
\begin{equation}
    \mathbb{E}_{\substack{x, x' \sim D\\ x\in \ls, x'\in \ds} }[\max\{\ell(f(x)) - \ell(f(x')),0\}].
\end{equation}
This definition measures the \emph{average} additional loss a model would experience for a random member of \ls versus \ds. While any number of measurements of ``fairness" of ML systems have been proposed (e.g., statistical parity, equality of false positives or negatives, calibration, individual fairness~\citep{hardt,fta,kleinberg,chouldechova2017fair}), many of them involve some per-instance loss function. Our measure does not perfectly capture all aspects of some model $f$'s behavior for all populations, but it does attempt to measure whether a model does a similarly good job minimizing a particular loss function for two different populations simultaneously.

%% file: evaluation.tex
\section{Evaluating Predictive Inequity}
Our goal is to quantify any disparity in predictive performance of standard recognition models across groups of people with varying skin tones. As no benchmark currently exists for this task, we first describe our methodology for collecting the necessary annotations for performing our evaluations in Section~\ref{sec:dataset}. Next, we provide evaluation of predictive inequity on our benchmark as well as an in depth analysis of the sources of inequity in Section~\ref{sec:eval}. Finally, we propose a simple remedy to reduce this predictive inequity in Section~\ref{sec:reduce_ineq}.

\input{dataset.tex}

\input{results_and_analysis.tex}
\input{reduce_inequity.tex}

%% file: dataset.tex
\subsection{Benchmarking Predictive Disparity for Pedestrian Detection}
\label{sec:dataset}

In order to measure predictive disparity of a particular model,  we need a partition of a dataset into demographic classes, in our case classes \ls and \ds. Many instances of bias in ML systems are those where elements of a dataset are not explicitly labeled by their demographic information. In these cases, we cannot necessarily assume the membership of data elements into \ls and \ds are known, and instead must gather that information as part of our training and evaluation of a system.

For the task of measuring the predictive inequity of object detection on pedestrians of different Fitzpatrick skin types, this corresponds to having a Fitzpatrick skin type label for each ground-truth pedestrian. Standard object detection datasets do not contain this information; given the size of these datasets, we decided to enlist the help of Mechanical Turk workers to categorize each pedestrian in the dataset with the information about their skin tone.

\subsubsection{A Tool for Collecting Annotations}

The vanilla BDD100K dataset lacks explicit labelings of people by skin color; each image instead is labeled by a set of bounding boxes along with a class label (the finest-grained class of pedestrians is the ``person" class). The dataset therefore needed to be augmented with the Fitzpatrick skin type of each pedestrian in order to measure various models' predictive inequity with respect to this categorization. We outline our approach to gathering these annotations below.

\begin{figure}
\begin{center}
\includegraphics[width=\linewidth]{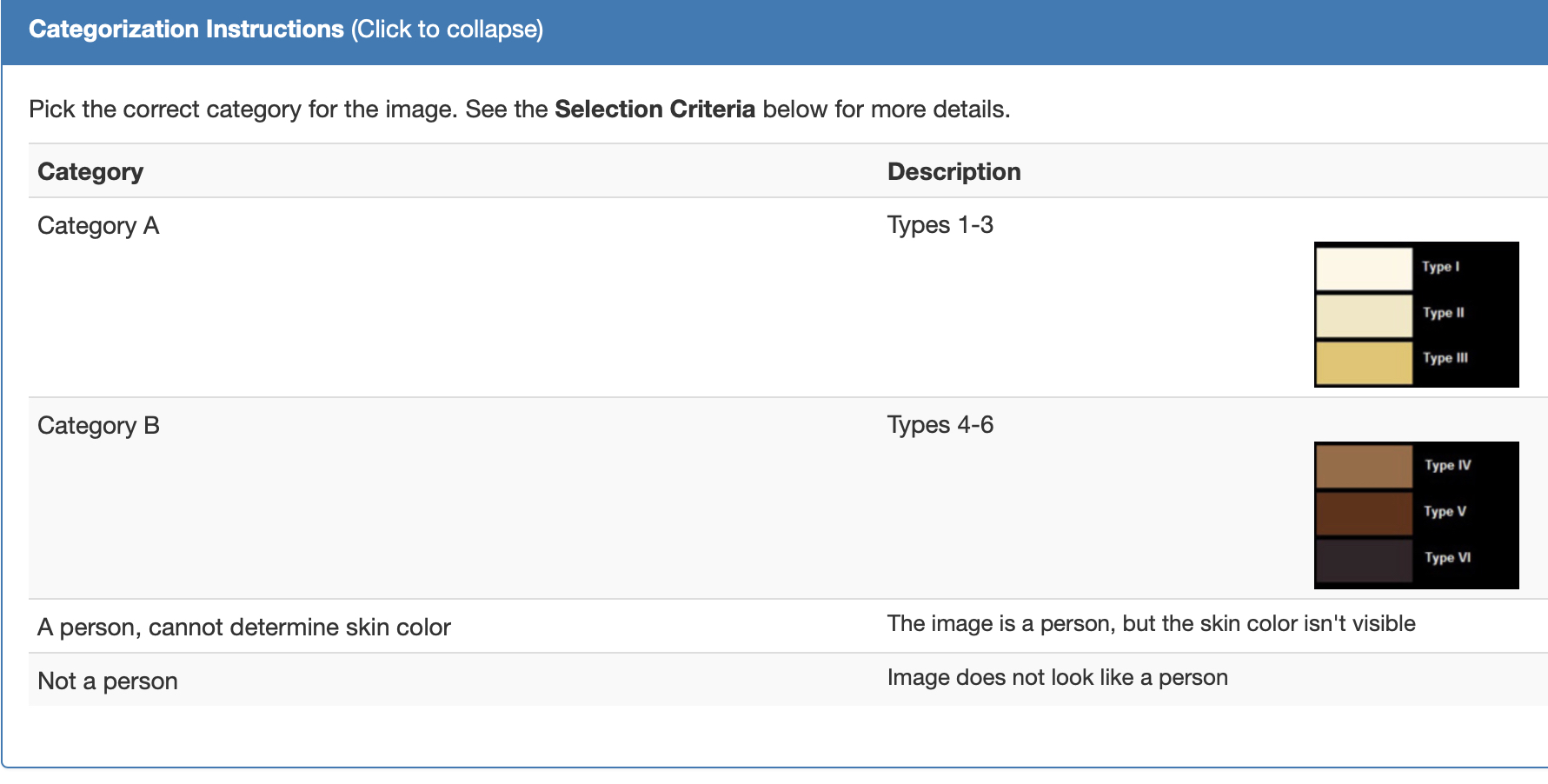}
\end{center}
\caption{Instructions given to mechanical turk annotators for classifying \ls and \ds people.}
\label{fig:annotation_tool}
\end{figure}

\begin{figure}
\begin{center}
\includegraphics[width=\linewidth]{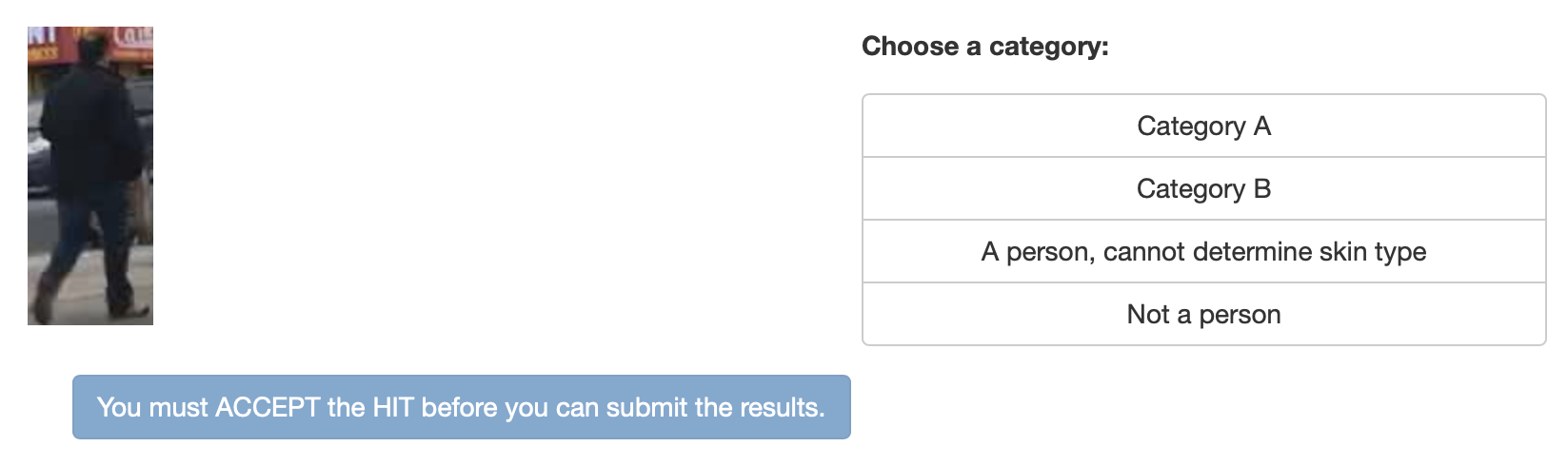}
\end{center}
\caption{Annotation interface.}
\label{fig:anno_interface}
\end{figure}

We initially cropped the bounding boxes of individuals that were labeled as the \textit{person} class. We then 
created tasks from each bounding box, asking for each pedestrian to be classified
into one of  4 categories: Fitzpatrick Categories 1-3, Fitzpatrick Categories 4-6, a person whose skin color cannot be determined, and not a person. The last two categories were included for multiple reasons: lighting, size, and occlusion can encumber determining the skin color of an individual with high confidence; moreover, BDD100K contains a small number of  mislabeled instances. The instructions presented to Turkers in shown in Figure~\ref{fig:annotation_tool} and an example annotation interface is shown in Figure~\ref{fig:anno_interface}.

Initially, we intended to use the entire set of pedestrians labeled within BDD100K; however, we quickly found issue with this process. We began by manually annotating a small random subset of these pedestrians, but found that even for the same annotator, there were substantial inconsistencies of the labels provided on the same instance when annotations were collected on separate occasions. This was due to many of the factors stated previously, but we attributed much of the disagreement to the extremely small size of many of the cropped images.
We found that this initial experiment had a large amount of disagreement, both amongst Turkers and compared to our labeling. We suspected this was due to the very small size of many of the cropped images.

We therefore focus our labeling and subsequent analysis on a  filtered set of cropped pedestrian images which contains only those individuals whose bounding box area was greater than or equal to 10,000 pixels, hoping to see an increased agreement among Turkers. With this cutoff, there were only 487 pedestrians in the validation class, so we labeled those ourselves. The  training set then had 4979 pedestrians above this cutoff, and for those we created Turker tasks. Due to the ongoing BDD100K challenge, \textit{person} bounding boxes for the testing set were not present; therefore, we did not acquire labels corresponding to skin color for the test dataset.
Imposing this minimum size constraint drastically increased the ease with which we were able to hand label the validation set, resulting in 3513 training images with a consensus of \ls or \ds labels, and 487 in validation.

\begin{figure}
    \centering
    \includegraphics[width=\linewidth]{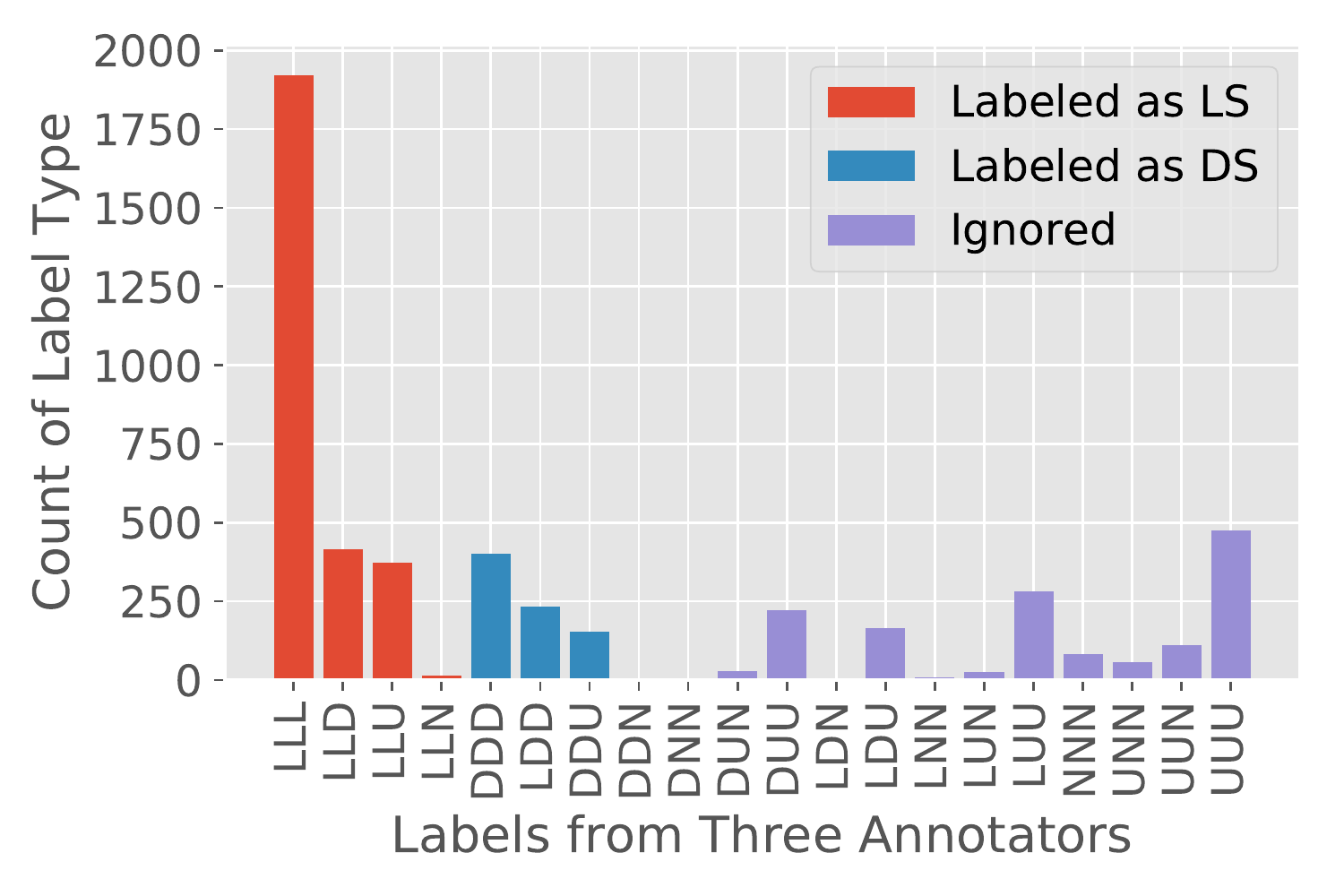}
    \caption{Histogram of the annotator responses. Each of the three annotators was given a choice of labeling as Category A (\ls -- denoted as \texttt{L}), Category B (\ds -- denoted as \texttt{D}), unknown (\texttt{U}), or not a person (\texttt{N}). Only instances with a consensus vote for \ls or \ds were labeled as such.} 
    \label{fig:hist_anno}

\end{figure}
For each person instance, we collected annotations from three separate Turkers. The set of possible labels we provided them with were: \ls, if the person could be clearly identified as from \ls; \ds if the person could be clearly identified as from \ds; \texttt{U}, if the skin color is unknown or too difficult to reliably identify; and \texttt{N}; if the box does not contain a person. A histogram of the received three score annotations received is shown in Figure~\ref{fig:hist_anno} and shows that the majority of identifiable people fall into \ls.

For each bounding box, if at least two of the three Turkers agreed on a label, we used that label; otherwise, we discarded that image for the purposes of evaluating predictive inequity. The summary of the number of consensus labelings can be found in Figure~\ref{fig:hist_anno}.

\begin{table}
    \centering
    \begin{tabular}{lcc}
    \toprule
    \textbf{Subset}     & \textbf{\ls} & \textbf{\ds} \\ 
    \midrule
    \textbf{Train}      & 2724       & 789        \\ 
    \textbf{Validation} & 387        & 100       \\
    \bottomrule
    \end{tabular}
    \caption{Count of labeled instances of people from \ls and \ds in BDD100K train and validation sets.}
    \label{tab:classbreakdown}
\end{table}

\subsubsection{Avoiding Annotator Bias}

Having humans annotate images of other humans based on skin tone opens up these annotations to scrutiny: will the labels be skewed based on cultural biases? Will they be accurate? How would we tell if the labels were skewed or inaccurate? For this reason and others, we inspected whether the distribution over categories \ls versus \ds from our hand labeling was similar to that of the distribution from the consensus labeling given by the aggregated Mechanical Turk annotations. We found that the rate of \ls to \ds was similar for both
(22\% from \ds based on Turker consensus on the training dataset and 20.5\% based on our hand labeling of the validation set). This suggests that the consensus labels from MTurk might have similar behavior to our hand labeling, and while this does not rule out bias that both systems share it does suggest some degree of precision between these methods.

In future work, we plan to do further validation between our hand labelings and MTurk labelings, and also consider different aggregation schemes for the Turker labels (Should the labels without universal agreement be labeled by more Turkers? Do our findings hold similarly for images with universal agreement on the skin type as they do on those with just $2$ agreements?).

\subsubsection{A Benchmark for Studying Predictive inequity}
The Berkeley Driving Dataset \citep{yu2018bdd100k} is one of the most comprehensive driving datasets to date. The dataset is comprised of both bounding box level and segmentation level annotations. The dataset includes 40 different classes common to driving scenes, and the images were taken from 4 different locations: New York, Berkeley, Downtown San Francisco, and the Bay Area near San Jose. Additionally, the dataset includes diverse weather conditions (rain, snow, sunshine), as well as images from various parts of the day (morning, dusk, and nighttime). The split for the bounding box level annotations is 70,000 images in the training set, 10,000 in the validation set, and 20,000 images in the test set, while the segmentation level annotations include 7,000 images in the training set, 2,000 images in the validation set, and 1,000 images in the test set. BDD100K is intended to be used to train real-time vision models that are included in current state of the art autonomous driving systems. Due to its relatively large size, we decided to use this dataset for experimentation.

%% file: results_and_analysis.tex
\subsection{Evaluating Predictive Inequity  of Standard Object Detection Systems}
\label{sec:eval}
In this section, we test whether several models displayed higher predictive inequity for pedestrians of Fitzpatrick types 4-6 as compared to those of types 1-3.
We begin by defining our evaluation setup as well as the metric used for quantitative evaluation. We evaluate object detection models on the task of recognizing people in the BDD100K validation set using the average precision metric. 

\paragraph{Metric: Average Precision}
A main metric used to quantify predictive performance of an object detection model is Average Precision (AP). For a given ground truth box, $b_\text{gt}$, and a predicted box, $b_\text{pred}$, the intersection over union of the predicted box is defined as the area of the intersection divided by the area of the union of the two boxes:
$$
\text{IoU} = \frac{b_{gt} \cap b_{\text{pred}}}{b_{gt} \cup b_{\text{pred}}}
$$
A predicted box is considered a true positive if it has IoU greater than a given threshold, $T$. Earlier challenges, such as Pascal~\citep{Everingham10}, focused on AP$_{50}$, which corresponds to a threshold, $T=0.5$. Current challenges, such as MS COCO~\citep{lin2014microsoft}, evaluate multiple metrics including their overall AP score which consists of averaging across thresholds in the range of $T\in[0.5,.95]$ by increments of $0.05$. Finally, the BDD100K challenge~\citep{yu2018bdd100k} from which we derive our benchmark focuses on a stricter localization goal of AP$_{75}$, corresponding to $T=0.75$. In the following sections we report performance for each of the three forms of evaluating average precision mentioned here.

\paragraph{Data Splits}
All of our results are reported on BDD100K validation images. When performance is measured for skin tones from \ls and \ds, we consider only the subset of the BDD100K validation images for which at least one person from either category is present. All people within an evaluation image which are too small to be reliably annotated into either \ls or \ds are ignored during per skin-type evaluation. We do this so as not to unnecessarily penalize the person detector for correctly predicting a small person.

\paragraph{Statistical validity}
We take a moment to describe how certain we are that the phenomena described below happen generally over a resampled validation set drawn from the same distribution as the validation set on which we report our empirical findings. In short, standard holdout-style arguments for a loss function of $k$ models evaluated on a holdout set of size $n$ will have $1-\delta$ probability of being within $\pm \sqrt{\frac{\ln\frac{k}{\delta}}{n}}$. If we view our validation set as a holdout, and consider only those bounding boxes for pedestrians of size at least 10,000 pixels (for which we have Fitzpatrick annotations), we only have holdout set sizes $n_\ls = 387, n_\ds = 100$. This means that our estimates on AP for category B for a single model are only accurate up to 0.17 for $\delta = .05$, and up to 0.08 for category A. So, if the true gap between AP for \ls versus \ds is .05, we would need $n_\ls = 12,000, n_\ds = 4,000$ to verify that \ls's AP surpassed that for \ds (with probability .95 over the draw of the holdout set).  

This suggests that gathering high-confidence comparisons between \ls and \ds will require much larger datasets (or, at least, many more examples of sufficiently large pedestrians for which we can gather Fitzpatrick annotations). We are not aware of any academic driving dataset with this many instances of large pedestrians, making it difficult without gathering a new dataset to validate our findings. 

However, we note that this behavior persists when evaluating on the larger training set and the validation set, in a fairly wide variety of learning settings, for a variety of models, throughout the training process, which should give some degree of confidence that this behavior is not entirely spurious. 

Moreover, we observed this behavior for  a wide range of learning rates (what generally is referred to  as a ``hyperparameter" of a learning algorithm); this phenomenon was not the result of picking a perfect training schedule.
We found consistently that models exhibited between 4 and 10\% gaps in  these precision metrics (with \ls consistently  outperforming \ds). For all models we train, we run each 10 times and report the mean and standard deviation in the resulting tables. \textbf{We emphasize  that the standard deviation listed in our experiments is solely a function of randomness in the training procedure and does not provide confidence intervals with respect to sampling error.}

\subsubsection{Training Data Importance}
We first compare performance for a state-of-the-art object detection model, Faster R-CNN~\citep{ren2015faster} model with a R-50-FPN~\citep{He2015} backbone, trained using two different sources of annotated data. We consider the standard learning protocol of initializing with ImageNet weights and then further training on a detection dataset. 

The MS COCO dataset~\citep{lin2014microsoft} is one of the predominant datasets currently used to evaluate an object detection model's performance, consisting of 80 different classes, including a \textit{person} class. Before studying predictive inequity, we begin by verifying the effectiveness of the person detector trained from MS COCO data for recognizing the pedestrians in the BDD100K validation dataset in Table~\ref{tab:bddval_compare_coco_vs_bdd}.
We compare this against the performance of the person detector trained using the BDD100K training set. We find that the MS COCO model outperforms the BDD100K model on the overall AP metric as well as the AP$_{50}$ metric, while the BDD100K model outperforms the MS COCO model on the strict localization evaluation of AP$_{75}$. Thus we find that both sources of training data are relevant for assessing person detection on this validation set. Implementation details and hyperparameters are available in the appendix.

\input{tables/BDDval_compare_coco_vs_bdd.tex}

\begin{figure}
    \centering
    \includegraphics[width=\linewidth]{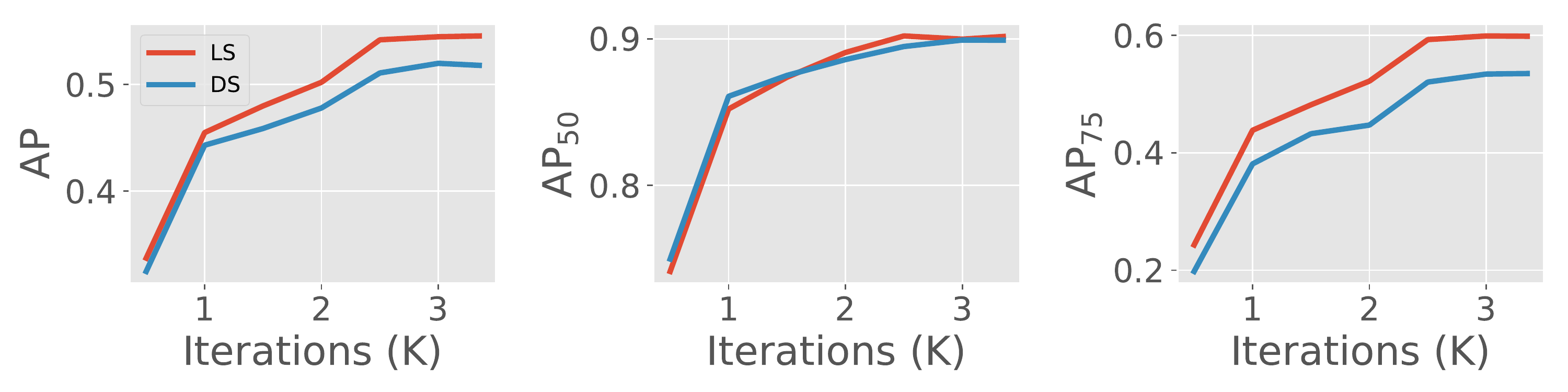}
    \caption{AP performance gap comparing \ls and \ds individuals using an unweighted model across training iterations on BDD100K. \ls consistently has higher AP then \ds people.}
    \label{fig:BDD_apgap_vs_iter}
\end{figure}

This leads us to our main question, do these models perform differently when evaluated on Fitzpatrick skin types [1-3] (\ls) vs skin types [4-6] (\ds)? To answer this question, we evaluate the same models as above as they perform at recognizing people identified within the BDD100K validation set (described in Section~\ref{sec:dataset}) annotated with \ls or \ds labels. Table~\ref{tab:bddval_compare_coco_vs_bdd_catAB} reports this breakdown. What we find is that consistently, across models trained with either MS COCO or BDD100K \emph{train}, people from \ls are recognized with higher average precision than people from \ds. The largest disparity occurs when evaluating using standard BDD100K metric of AP$_75$, which requires tight localization.
\input{tables/BDDval_compare_coco_vs_bdd_catAB.tex}  We see a much higher level of predictive inequity using MS COCO weights compared to BDD100k weights on AP$_{75}$, suggesting this problem is not unique to the BDD100k training data.
We also verify that this discrepancy between AP on \ls vs. \ds is not an artifact of a particular point in training and instead persists across training iterations as shown in Figure~\ref{fig:BDD_apgap_vs_iter}.

\begin{remark}
 MS COCO contains a broad set of classes, such as ``umbrella" and ``suitcase", which are generally not included within a ``person" bounding box, while BDD100k's ``person" bounding box often includes these (making the ground truth annotations somewhat different).
\end{remark}

\begin{figure*}[t]
    \centering
    \includegraphics[width=.32\linewidth]{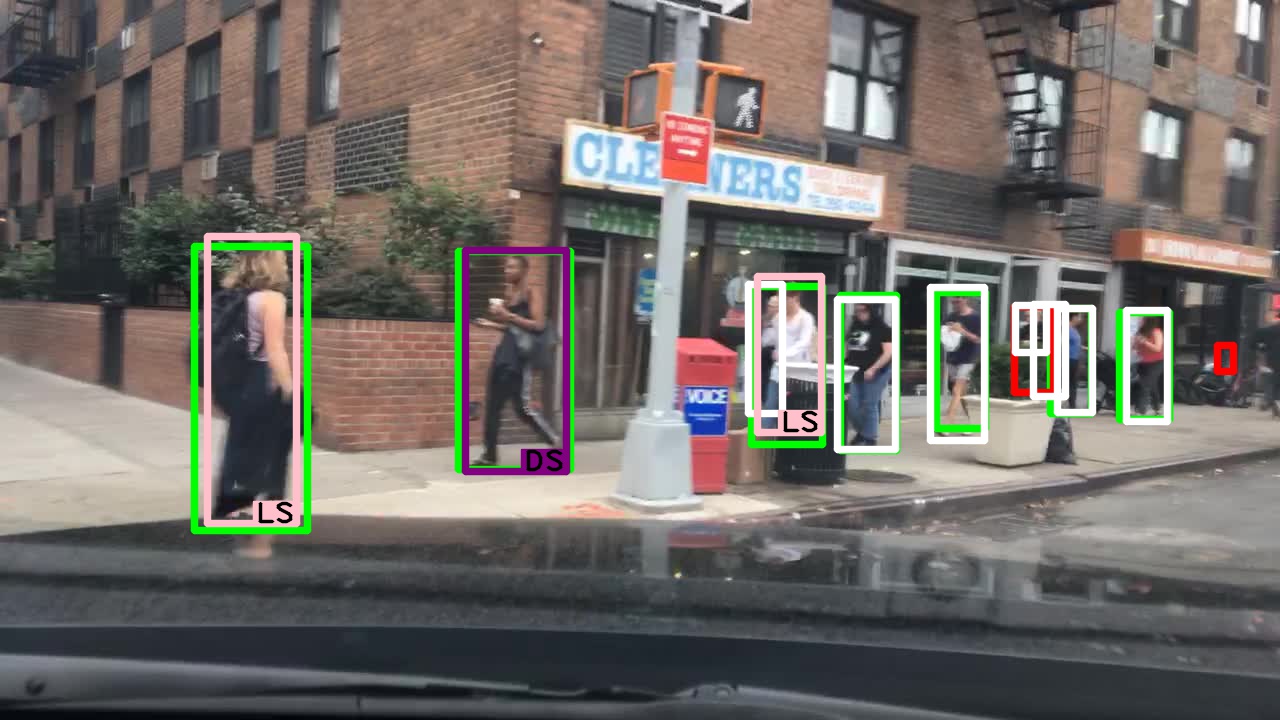}
    \hfill
    \includegraphics[width=.32\linewidth]{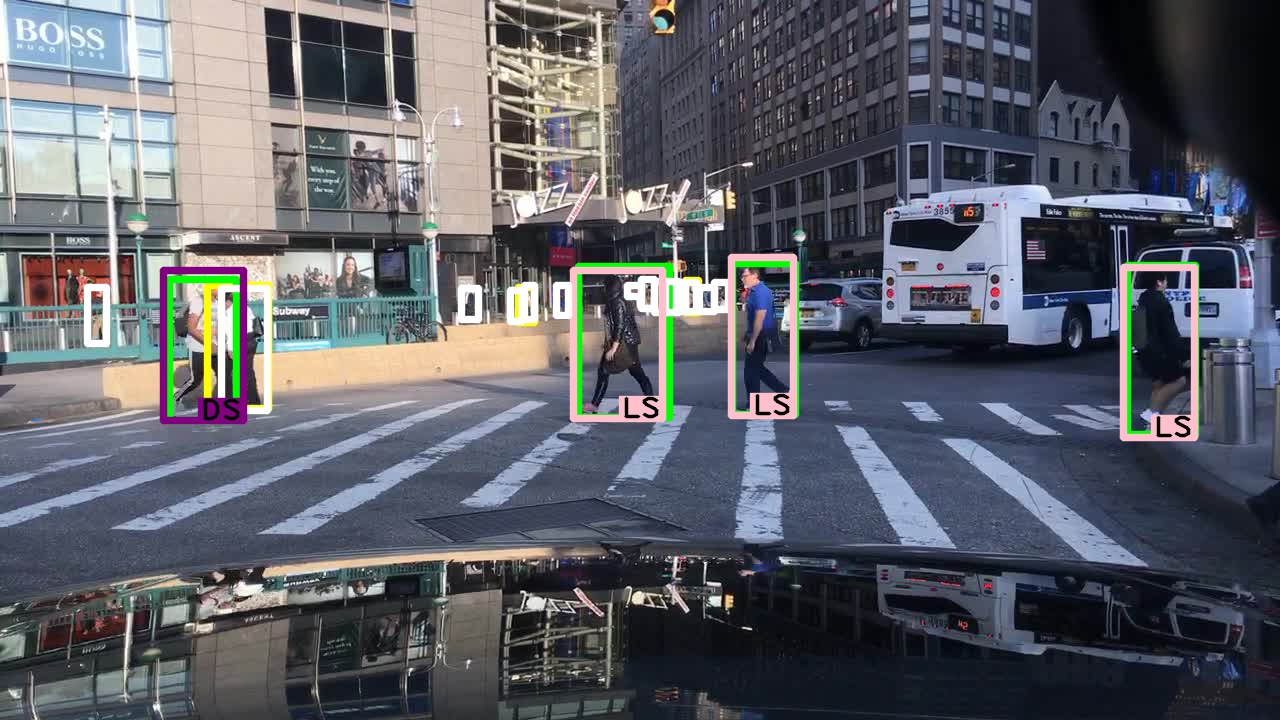}
    \hfill
    \includegraphics[width=.32\linewidth]{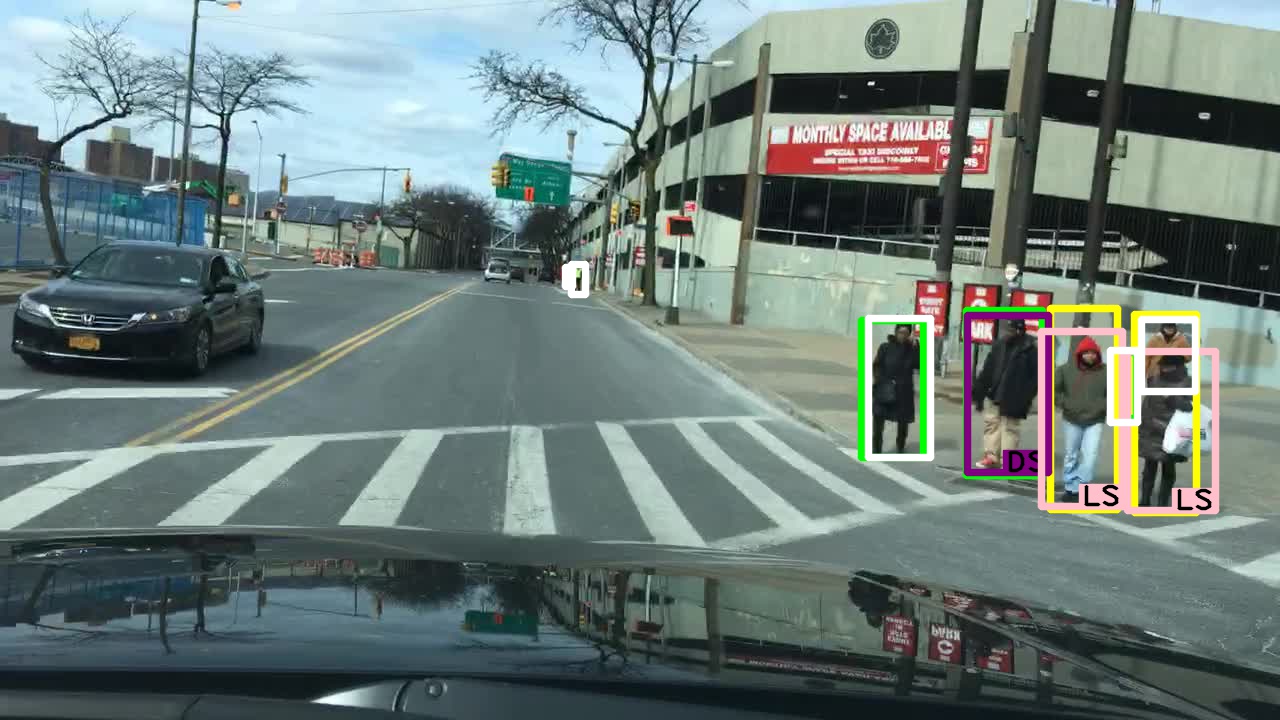}
    \hfill
    \includegraphics[width=.32\linewidth]{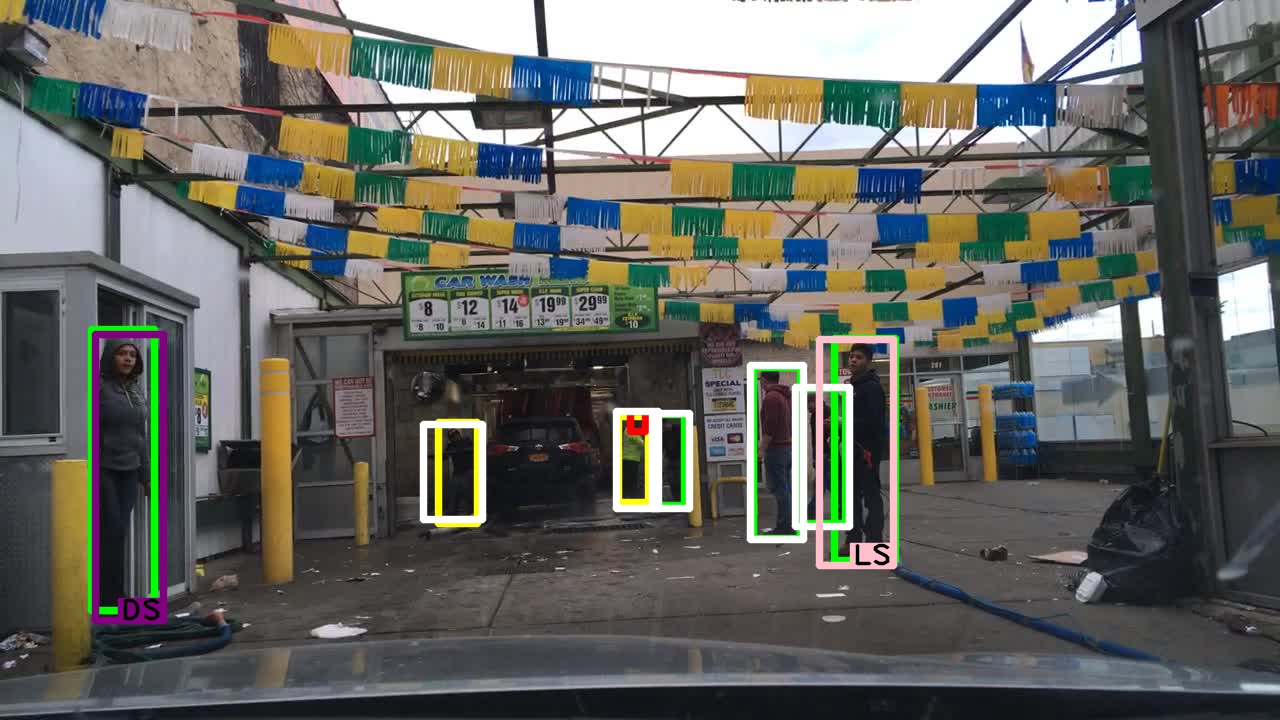}
    \hfill
    \includegraphics[width=.32\linewidth]{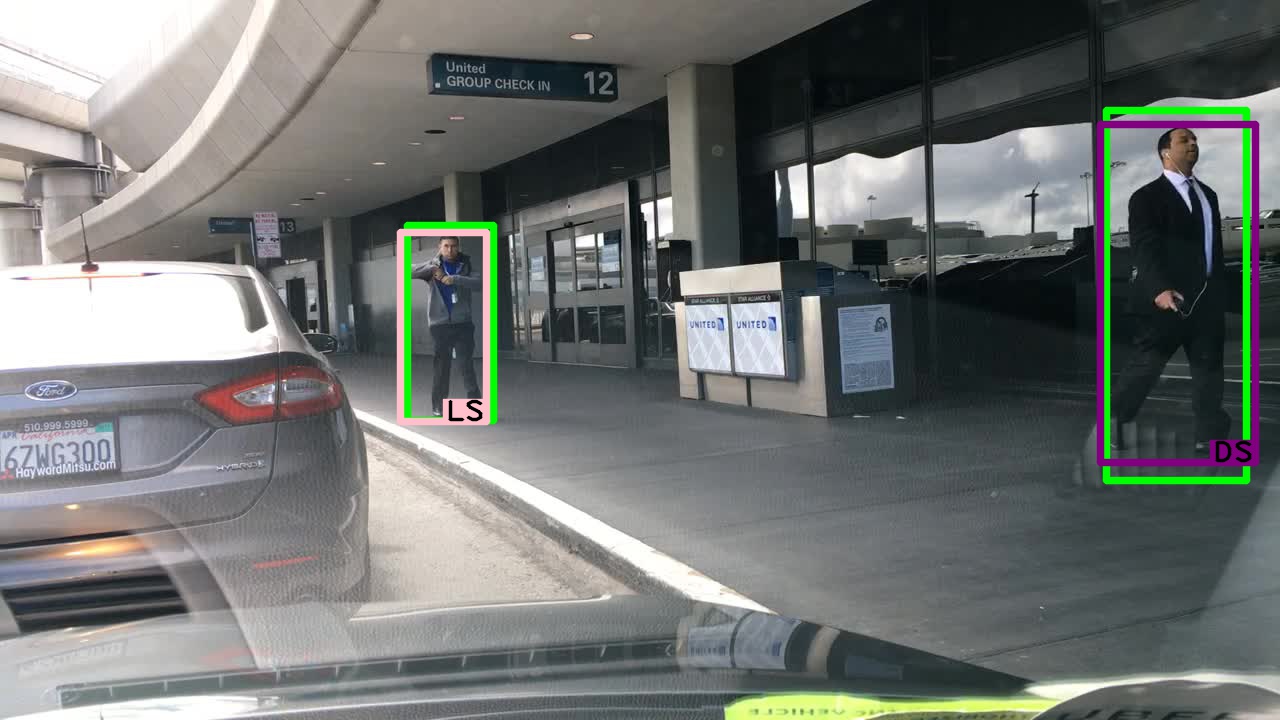}
    \hfill
    \includegraphics[width=.32\linewidth]{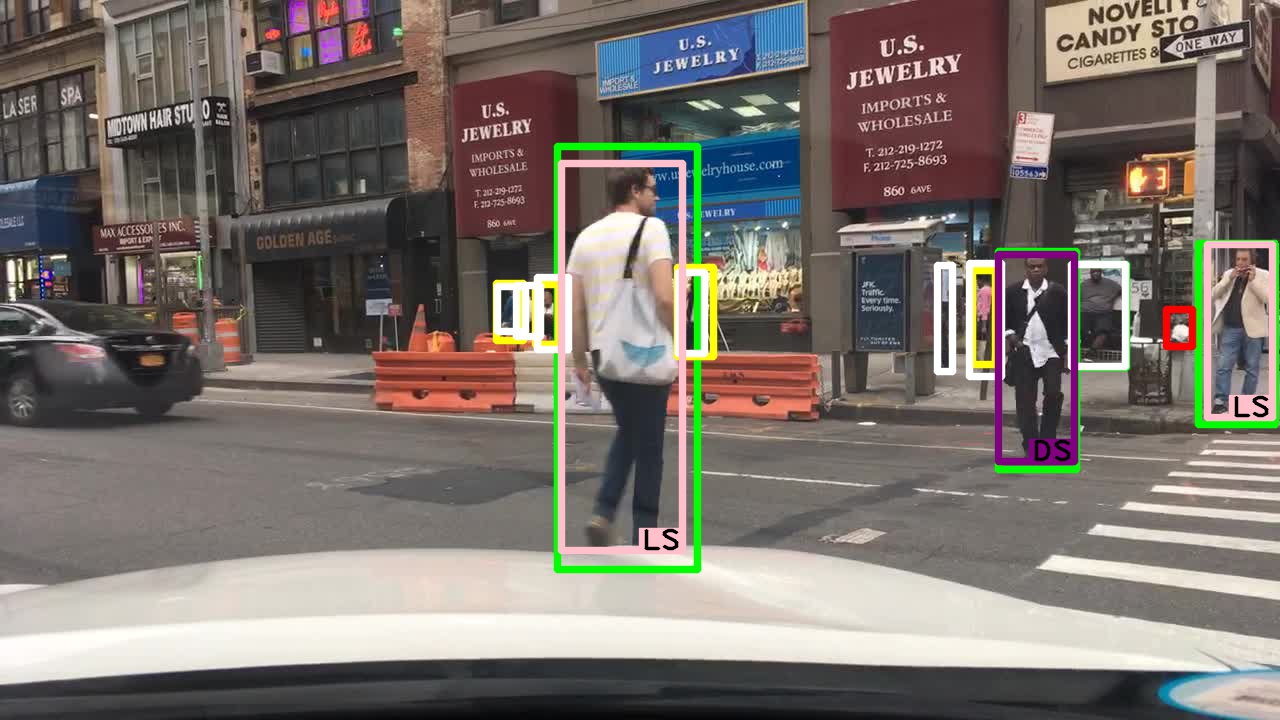}
    \hfill
    \caption{Example detections from Faster R-CNN using the R-50-FPN backbone, trained on BDD100K. For reference, the ground truth annotations for \ls and \ds are pink and purple respectively. Yellow boxes correspond to true positives under the AP$_{50}$ metric and false positives under the AP$_{75}$ metric. Green boxes correspond to true positives under the AP$_{75}$ metric. All the predictions shown are greater than an 85\% confidence threshold.}
    \label{fig:ExampleDetectionsCorrect}
\end{figure*}

\subsubsection{Importance of Architecture and Model Selection}
Until now, we have shown evidence of predictive inequity  between people of skin tones in the Fitzpatrick range [1-3] (\ls) as compared to range [4-6] (\ds) for a single object detection model learned using two different sources of data. A reasonable question to ask is whether or not this observation is an artifact of that particular architecture or model. Therefore we study prediction of people across the two skin-type categories across multiple architecture backbones and two state-of-the-art models, Faster R-CNN~\citep{ren2015faster} and Mask R-CNN~\citep{he2017mask}. We evaluate the publicly available model weights from training on the MS COCO dataset, released within the Detectron model zoo~\citep{Detectron2018} and report performance in Table~\ref{tab:bddval_mscoco}.

\input{tables/BDDval_MSCOCO_weights.tex}

We find that across all models and base architectures studied, performance on \ls exceeds that of \ds, demonstrating that this phenomenon is not specific to a particular model.

Again, the most striking measure of predictive inequity is observed under the strict localization metric of AP$_{75}$, where the average performance across models studied drops from 67.1\% for \ls to 55.9\% for \ds. However, under the weaker localization metric of AP$_{50}$, the gap between the two groupings of people is greatly diminished. In the next section we provide further analysis into the results shown here to explain the discrepancy between the two metrics.   

\subsection{Analyzing Sources of Predictive inequity }

\input{tables/BDDval_MSCOCO_no_occluded_weights.tex}

In the previous section we demonstrated that using the tight localization metric of AP$_{75}$ a wide variety of object detection models and training schemes results in predictive inequity  between individuals with skin tones from \ls and \ds. Here we investigate the natural followup question: what causes this discrepancy?

We begin our study by inspecting example output detections in Figure~\ref{fig:ExampleDetectionsCorrect}. We show here ground truth (white box) with \ls or \ds indicated where appropriate for people appearing in each image. We then show all boxes with scores greater than a threshold (0.85). We indicate those boxes which are false positives under all metrics in red. Those which have overlap $\geq 0.5$ with the ground truth, but $<0.75$ and are thus false positives under the AP$_{75}$ metric are indicated in yellow. Finally, those which are true positives under all metrics are shown in green. 

By studying such images, we observe potential sources of predictive inequality which we will analyze next in more detail; namely, occlusion, time of day, and loss prioritization. 

\subsubsection{Occlusion as a possible source of predictive inequity }

One observation is that often street scenes contain multiple people on the sidewalk together or crossing the road near each other leading to partially occluded individuals. Recognition under occlusion is known to be more challenging (see Figure~\ref{fig:ExampleDetectionsCorrect} top right 
and if the dataset has a bias by which one type of individual occurs more frequently in crowded scenes with occlusion, it would be unsurprising if that type of individual suffered lower performance in overall recognition.

To remove this confounding factor we study the performance of a reduced ground truth set which contains only the unoccluded people from BDD100K validation for which we have discernible skin type labels. Table~\ref{tab:bddval_mscoco_no_occluded} reports the three average precision metrics and results in consistent observations with those made in the full validation set which included the occluded individuals. We observe that AP, AP$_{50}$, and AP$_{75}$ all appear to improve for both \ls and \ds once these
occluded examples are removed, but that the gap between \ls and \ds performance for AP and AP$_{75}$ remains.  Therefore, we conclude that the source of discrepancy between performance on \ls and \ds is not due to co-occurrence with occluded people.\footnote{We remark that considering performance broken down by occlusion was actually motivated by unexpected performance of these MSCOCO weighted models on BDD100K training data; see section~\ref{sec:train} for further discussion.}

\subsubsection{Time of Day as a Possible Source of Predictive Inequity}

Low contrast between a subject and its background may affect the recognition performance of that subject. In outdoor street scenes the natural lighting variations that occur at various times of day will change the contrast of the person itself and between the person and the background, potentially introducing a confounding factor when comparing predictive performance between \ls and \ds. Therefore, we next measure whether the observed gap in predictive performance is attributable to time of day by reporting performance during daytime or nighttime hours, annotations which are available within BDD100K.

Table~\ref{tab:daytime-vs-nighttime} reports the recognition performance for individuals from \ls and \ds who appear in day versus night. We used Faster R-CNN with R-50-FPN backbone trained on BDD100K \emph{train} with $\alpha_{\ds}$ = 1 for evaluation. Focusing only on daytime images, for each of AP, AP$_{50}$, and AP$_{75}$, performance of our trained model on the \ls category was higher than that on the \ds category, and 
the size of the gap in each case was quite similar to that on the entire validation set. Surprisingly, for nighttime images, the performance of our model on \ds pedestrians was higher than on \ls pedestrians. However, we note that this may be the result of small nighttime sample size (the validation set has only 144 nighttime pedestrians of sufficiently large size, only 15 of which are designated as \ds).  Nonetheless, this data suggests an important conclusion that \emph{there is not evidence that time of day is to blame for the predictive inequity observed overall.}

\subsubsection{Loss Prioritization as a possible source of predictive inequity }
We know that there are more then three times as many individuals from \ls as from \ds in the BDD100K dataset (see Table~\ref{tab:classbreakdown}). In the next section we take steps to reduce this higher degree of representation for \ls in the training set 
 through supervised loss weightings. This will give some indication as to what fraction of the observed  predictive inequity stems from prioritizing loss on pedestrians from \ls over those from \ds, and whether a simple solution (namely, reweighting the loss function) can effectively treat some fraction of this observed predictive inequity. 

%% file: tables/BDDval_compare_coco_vs_bdd.tex
\begin{table}
    \begin{center}
    \resizebox{\linewidth}{!}{
    \begin{tabular}{l  c c c}
    	\toprule
    	  \textbf{Training Data}   & \textbf{AP (\%)}& \textbf{AP$_{50}$ (\%)} & \textbf{AP$_{75}$ (\%)}\\
    	\midrule
    	 \text{MS COCO}&   24.2 & 56.9 & 16.0 \\
    	 \text{BDD100K \emph{Train}}& 22.3 $\pm$ 2.0 & 45.4 $\pm$ 5.2 & 18.8 $\pm$ 1.0 \\
    	 \bottomrule
    \end{tabular}
    }
    \end{center}
    \caption{Average precision of the person class on BDD100K \emph{validation}. Performance here is across all people.}
    \label{tab:bddval_compare_coco_vs_bdd}
    
\end{table}

%% file: tables/BDDval_compare_coco_vs_bdd_catAB.tex
\begin{table}
    \begin{center}
    \resizebox{\linewidth}{!}{
    \begin{tabular}{ll  ccc ccc ccc}
    	\toprule
    	  \multirow{2}{*}{\textbf{Training Data}}  && \multicolumn{2}{c}{\textbf{AP (\%)}}       && \multicolumn{2}{c}{\textbf{AP$_{50}$ (\%)}} && \multicolumn{2}{c}{\textbf{AP$_{75}$ (\%)}}\\
    	  \cmidrule{3-4}\cmidrule{6-7}\cmidrule{9-10}
     	  & & \ls & \ds && \ls & \ds && \ls & \ds \\
    	\midrule
    	 \text{MS COCO}&&   \bf 57.2 & 53.3 && 91.3 & \bf 91.8 && \bf 63.1 & 53.6 \\
    	 \text{BDD100K \emph{Train}} && \textbf{54.6 $\pm$ 0.4} & 51.8 $\pm$ 0.9 && \textbf{90.2 $\pm$ 0.9} & 89.9 $\pm$ 1.5 && \textbf{59.8 $\pm$ 1.0} & 53.5 $\pm$ 1.7 \\
    	 \bottomrule
    \end{tabular}
    }
    \end{center}
    
    \caption{Comparing average precision on BDD100K \emph{validation} set across the larger people labeled as \ls or \ds for Faster-RCNN R-50-FPN trained either with MS COCO or BDD100K train data (averaged over 10 different trainings).}
    \label{tab:bddval_compare_coco_vs_bdd_catAB}
    
\end{table}

%% file: tables/BDDval_MSCOCO_weights.tex
\begin{table}[t]
    \begin{center}
    \resizebox{\linewidth}{!}{
    \begin{tabular}{l l  cc c cc c cc}
    	\toprule
    	  \multirow{2}{*}{\textbf{Model}} & \multirow{2}{*}{\textbf{Backbone}}        & \multicolumn{2}{c}{\textbf{AP (\%)}}       && \multicolumn{2}{c}{\textbf{AP$_{50}$ (\%)}} && \multicolumn{2}{c}{\textbf{AP$_{75}$ (\%)}}\\
    	  \cmidrule{3-4}\cmidrule{6-7}\cmidrule{9-10}
     	  & & \ls & \ds && \ls & \ds && \ls & \ds \\
    \hline
     \textbf{Faster R-CNN} & \textbf{R-50-C4}         & \textbf{57.3} & 52.7 && \textbf{90.3} & 90.0 && \textbf{64.5} & 53.6                \\
     \textbf{Faster R-CNN} & \textbf{R-50-FPN}        & \textbf{57.2} & 53.3 && 91.3 & \textbf{91.8} && \textbf{63.1} & 53.6                \\
     \textbf{Faster R-CNN} & \textbf{R-101-FPN}       & \textbf{59.7} & 56.9 && 91.6 & \textbf{93.0} && \textbf{70.0} & 57.3                \\
     \textbf{Faster R-CNN} & \textbf{X-101-32x8d-FPN} & \textbf{60.6} & 55.4 && \textbf{93.1} & 90.0 && \textbf{69.8} & 62.1                \\
     \textbf{Mask R-CNN}   & \textbf{R-50-C4}         & \textbf{59.3} & 53.8 && \textbf{91.3} & 90.5 && \textbf{66.9} & 51.5                \\
     \textbf{Mask R-CNN}   & \textbf{R-50-FPN}        & \textbf{58.9} & 53.2 && \textbf{92.5} & 92.3 && \textbf{67.6} & 51.2                \\
     \textbf{Mask R-CNN}   & \textbf{R-101-FPN}       & \textbf{60.1} & 54.7 && \textbf{92.4} & 92.1 && \textbf{65.6} & 55.2                \\
     \textbf{Mask R-CNN}   & \textbf{X-101-32x8d-FPN} & \textbf{60.8} & 57.1 && \textbf{93.3} & 91.6 && \textbf{69.2} & 62.9                \\
     \hline
     \multicolumn{2}{c}{\textbf{Average}} & \textbf{59.2}         & 54.6         && \textbf{92.0}         & 91.4         && \textbf{67.1}         & 55.9         \\
    \hline
    \end{tabular}
    }
    \end{center}
    \caption{Performances of the person class on BDD100K \emph{validation} set for models trained using MS COCO. We note that these models were not trained on BDD100K.}
    \label{tab:bddval_mscoco}
    
\end{table}

%% file: tables/BDDval_MSCOCO_no_occluded_weights.tex
\begin{table}
    \begin{center}
    \resizebox{\linewidth}{!}{
    \begin{tabular}{l l  cc c cc c cc}
    	\toprule
    	  \multirow{2}{*}{\textbf{Model}} & \multirow{2}{*}{\textbf{Backbone}}        & \multicolumn{2}{c}{\textbf{AP (\%)}}       && \multicolumn{2}{c}{\textbf{AP$_{50}$ (\%)}} && \multicolumn{2}{c}{\textbf{AP$_{75}$ (\%)}}\\
    	  \cmidrule{3-4}\cmidrule{6-7}\cmidrule{9-10}
     	  & & \ls & \ds && \ls & \ds && \ls & \ds \\
    \hline
     \textbf{Faster R-CNN} & \textbf{R-50-C4}         & \textbf{62.0} & 57.1 && 93.5 & \textbf{94.1} && \textbf{72.9} & 60.7                \\
     \textbf{Faster R-CNN} & \textbf{R-50-FPN}        & \textbf{61.2} & 58.0 && 93.5 & \textbf{95.8} && \textbf{70.9} & 59.4                \\
     \textbf{Faster R-CNN} & \textbf{R-101-FPN}       & \textbf{64.2} & 60.8 && \textbf{95.7} & 95.3 && \textbf{77.0} & 63.2                \\
     \textbf{Faster R-CNN} & \textbf{X-101-32x8d-FPN} & \textbf{64.7} & 59.4 && \textbf{95.7} & 92.8 && \textbf{75.9} & 68.9                \\
     \textbf{Mask R-CNN}   & \textbf{R-50-C4}         & \textbf{63.4} & 58.2 && 94.0 & \textbf{95.0} && \textbf{74.1} & 58.7                \\
     \textbf{Mask R-CNN}   & \textbf{R-50-FPN}        & \textbf{63.2} & 56.3 && \textbf{95.3} & 94.7 && \textbf{75.6} & 56.1                \\
     \textbf{Mask R-CNN}   & \textbf{R-101-FPN}       & \textbf{64.3} & 58.5 && 95.3 & \textbf{95.5} && \textbf{73.5} & 59.1                \\
     \textbf{Mask R-CNN}   & \textbf{X-101-32x8d-FPN} & \textbf{64.9} & 61.4 && \textbf{96.1} & 95.0 && \textbf{76.6} & 70.5                \\
     \hline
     \multicolumn{2}{c}{\textbf{Average}} & \textbf{63.5}           & 58.7           && \textbf{94.9}           & 94.8           && \textbf{74.6}           & 62.1           \\
    \hline
    \end{tabular}
    }
    \end{center}
    \caption{Average precision on BDD100K validation set with occluded individuals removed for models trained using MS COCO.
    }
    \label{tab:bddval_mscoco_no_occluded}
\end{table}

%% file: reduce_inequity.tex
\section{Reducing Predictive Inequity}
\label{sec:reduce_ineq}

\input{tables/BDDval_BDDtrain_compare-weightings.tex}

\newcommand{\ce}{\textrm{cls}}

\newcommand{\reg}{\textrm{1-reg}}
We now describe our investigation into whether the importance of \ls's loss 
compared to that of \ds's loss during training is a source of the predictive inequity between the classes. Many loss functions (including those regularly used to train the models studied in this work) decompose into terms for each training example, weighting each example uniformly. This tacit weighting implies that a subset of the training data which is 3.5 times larger than another subset (as \ls is compared to \ds) will have up to 3.5 times as much impact on such a loss function, possibly steering an algorithm towards models which have lower loss on \ls compared to \ds.  We emphasize here that such issues are not information-theoretic: such issues stem from the design choice of loss function rather than having too little data to give generalizable predictions for the smaller subset.

For the purpose of this discussion we will describe our methodology in terms of a generic loss function (see Appendix for the longer and more precise discussion specific to Faster R-CNN), $\mathcal{L}(x,y)$, which takes as input an image crop, $x$, and the true label, $y$, which in our case will be either person, some other class, or background. Let us indicate the set of people instances which are also labeled as \ls or \ds as $X_\ls$ and $X_\ds$ respectively, with the number of instances denoted as $N_\ls$ and $N_\ds$ respectively. Let all other boxes not containing a person from \ls or \ds be denoted as $X_O$. Then we can define our overall loss in terms of a weighted sum over each instance:
\begin{eqnarray}
    \text{Total Loss} &= \frac{\alpha_{\ls}}{N_\ls} \sum_{\{x_\ls, y_\ls\} \in X_\ls} \mathcal{L}(x_\ls, y_\ls)\\
        & + \frac{\alpha_\ds}{N_{\ds}} \sum_{\{x_\ds, y_\ds\} \in X_\ds} \mathcal{L}(x_\ds, y_\ds)\nonumber \\
        & + \frac{\alpha_O}{N_O} \sum_{\{x_o, y_o\} \in X_O} \mathcal{L}(x_o, y_o) \nonumber 
\end{eqnarray}
where $\alpha_\ls, \alpha_\ds, \alpha_O$ denotes the per class weighting on instances from \ls, \ds, or other, respectively. 

To measure whether our loss function emphasizes accuracy for \ls compared to \ds and in so doing creates some predictive inequity, we consider several reweightings of the standard loss function used in training Faster R-CNN. The (standard) unweighted or equal weighted loss function for Faster R-CNN has two components, one aimed at affecting the region proposal network and the other for the detection and classification inside these proposed regions. We only consider weightings which affect the latter part of the loss function. 

The detection and classification component of the Faster R-CNN loss function combines a cross-entropy term $\mathcal{L}_{\ce}$
and a regularized $\ell_1$ loss $\mathcal{L}_{\text{reg}}$ term 
for each example. For the purpose of reweighting, we consider the loss to be the joint cross-entropy and regularized terms and consider variants of values for $\alpha_\ls, \alpha_\ds$, and $\alpha_O$. To reduce the number of hyperparameters to optimize we fix both $\alpha_\ds$ and $\alpha_O$ to be 1 for all experiments and consider the effect of placing greater weight on \ds instances through raising $\alpha_\ds$.  

Average precision as we vary $\alpha_\ds$ weight on instances from \ds are reported in Table~\ref{tab:bddval_compare-weighting3}.
We find that the gap between the total AP value is reduced for larger values of $\alpha_\ds$, but that the \ls to \ds AP$_{75}$ gap is quite similar for each of the weightings, in the range of 4 to 6\%. For some of these weightings, in particular for $\alpha_\ds =3 $, we notice that the AP$_{75}$ for \ds pedestrians is quite close to the AP$_{75}$ for \ls pedestrians trained on unweighted examples. Moreover, the
performance of the model on \ls pedestrians is \emph{better} under this weighting.

This result suggests that careful reweighting may  improve performance on \ds pedestrians without sacrificing performance on \ls pedestrians, or even improve performance on \ls pedestrians as a byproduct.  Further analysis is needed to fully understand the effect of re-weighting on the stricter criteria used as part of the total AP metric to fully reduce the inequity. Overall, this finding suggests that predictive inequity stemming from the oft-labeled ``too little data" source might actually confound two fundamentally different phenomena: that smaller datasets beget less statistical certainty, but also tend to receive lower emphasis during training.

\input{tables/BDDval_daytime_vs_nighttime.tex}

%% file: tables/BDDval_BDDtrain_compare-weightings.tex
\begin{table}
    \begin{center}
    \resizebox{\linewidth}{!}{
    \begin{tabular}{l  cc c cc c cc}
    	\toprule
    	  \multirow{2}{*}{\textbf{$\alpha_\ds$}}   & \multicolumn{2}{c}{\textbf{AP (\%)}}       && \multicolumn{2}{c}{\textbf{AP$_{50}$ (\%)}} && \multicolumn{2}{c}{\textbf{AP$_{75}$ (\%)}}\\
    	  \cmidrule{2-3}\cmidrule{5-6}\cmidrule{8-9}
    	  & \ls & \ds && \ls & \ds && \ls & \ds \\
    \midrule
     1 & \textbf{54.6} $\pm$ \textbf{0.4} & 51.8 $\pm$ 0.9 && \textbf{90.2} $\pm$ \textbf{0.9} & 89.9 $\pm$ 1.5 && \textbf{59.8} $\pm$ \textbf{1.0} & 53.5 $\pm$ 1.7                \\
     2 & \textbf{55.3} $\pm$ \textbf{0.7} & 53.3 $\pm$ 0.6 && \textbf{90.9} $\pm$ \textbf{0.6} & 90.4 $\pm$ 1.1 && \textbf{60.8} $\pm$ \textbf{1.7} & 55.7 $\pm$ 3.3                \\
     3 & \textbf{55.8} $\pm$ \textbf{0.6} & 53.9 $\pm$ 1.3 && \textbf{91.3} $\pm$ \textbf{0.7} & 90.6 $\pm$ 1.0 && \textbf{63.4} $\pm$ \textbf{1.8} & 58.3 $\pm$ 2.9                \\
     5 & \textbf{56.4} $\pm$ \textbf{0.4} & 53.9 $\pm$ 1.1 && \textbf{91.8} $\pm$ \textbf{0.5} & 90.8 $\pm$ 1.0 && \textbf{63.0} $\pm$ \textbf{1.9} & 57.3 $\pm$ 2.3                \\
     10 & \textbf{55.8} $\pm$ \textbf{0.8} & 54.0 $\pm$ 1.0 && \textbf{91.7} $\pm$ \textbf{0.9} & 91.0 $\pm$ 0.8 && \textbf{62.1} $\pm$ \textbf{1.5} & 56.3 $\pm$ 2.0                \\
    \bottomrule
    \end{tabular}
    }
    \end{center}
    \caption{Performances of Faster-RCNN using R-50-FPN backbone on BDD100k \emph{validation} with different weightings on \ds in the classification network loss function. }
    \label{tab:bddval_compare-weighting3}
\end{table}

%% file: tables/BDDval_daytime_vs_nighttime.tex
\begin{table}
    \begin{center}
    \resizebox{\linewidth}{!}{
    \begin{tabular}{l cc c cc c cc}
    	\toprule
    	 \multirow{2}{*}{\textbf{Time}}   &
    	 \multicolumn{2}{c}{\textbf{AP (\%)}}       && \multicolumn{2}{c}{\textbf{AP$_{50}$ (\%)}} && \multicolumn{2}{c}{\textbf{AP$_{75}$ (\%)}}\\
    	  \cmidrule{2-3}\cmidrule{5-6}\cmidrule{8-9}
    	  & \ls & \ds && \ls & \ds && \ls & \ds \\
    \midrule
        Day & \textbf{57.2 $\pm$ 0.5} & 54.6 $\pm$ 1.2 && \textbf{91.8 $\pm$ 0.8} & 91.0 $\pm$ 1.7 && \textbf{63.9 $\pm$ 1.6} & 58.4 $\pm$ 2.2 \\
        Night & 43.3 $\pm$ 0.9 & \textbf{50.8 $\pm$ 1.6} && 81.4 $\pm$ 2.4 & \textbf{91.7 $\pm$ 3.7} && 41.7 $\pm$ 3.0 & \textbf{53.2 $\pm$ 5.0} \\
    \bottomrule
    \end{tabular}
    }
    \end{center}
    \caption{Performances at different times of day using Faster R-CNN with R-50-FPN backbone on BDD100K \emph{validation} using weights trained on BDD100K \emph{train} with $\alpha_{\ds}$ = 1. Note the number of daytime and nighttime examples are as follows: Day: \ls 297 \ds 75, Night: \ls 69 \ds 15.}
    \label{tab:daytime-vs-nighttime}
\end{table}

%% file: conclusion.tex
\section{Conclusion and Discussion}
\label{sec:conclusion}

In this work, we propose the concept of predictive inequity in detecting pedestrians of different skin tones 
in object detection systems. We give evidence that standard models for the task of object detection, trained on standard datasets, appear to exhibit higher precision on lower Fitzpatrick skin types than higher skin types. 
This behavior appears on large images of pedestrians, and even grows when we remove occluded pedestrians. Both of these cases (small pedestrians and occluded pedestrians) are known difficult cases for object detectors, so even on the relatively ``easy" subset of pedestrian examples, we observe this predictive inequity.  We have shown that simple changes during learning (namely, reweighting the terms in the loss function) can partially mitigate this disparity. We hope this study provides compelling evidence of the real problem that may arise if this source of capture bias is not considered before deploying these sort of recognition models.

%% file: appendix.tex
\newpage
\pagebreak
\clearpage
\section{Appendix}\label{sec:appendix}

\subsection{MSCOCO weighted models and occlusion}\label{sec:train}

We also evaluated the MSCOCO weighted models on the BDD100k training set,
primarily for the following reason. Because these model were not trained on BDD100K,
one can make stronger statistical claims about results of these models evaluated o
n this larger set (treating it as validation).  Interestingly, we found the gap
we observed on the validation set did not seem to persist on the larger training set---this presumably meant that either the results on the validation set were the result of sampling error, or there was some large statistical difference between the train/val sets for BDD100k. Upon further inspection, we discovered that occluded pedestrians affected these results significantly. We show the results of these experiments below in Tables~\ref{tab:bddtrain_mscoco_weights} and ~\ref{tab:bddtrain_mscoco_no_occluded}.

\begin{table}
    \begin{center}
    \resizebox{\linewidth}{!}{
    \begin{tabular}{l l  cc c cc c cc}
    	\toprule
    	  \multirow{2}{*}{\textbf{Model}} & \multirow{2}{*}{\textbf{Backbone}}        & \multicolumn{2}{c}{\textbf{AP (\%)}}       && \multicolumn{2}{c}{\textbf{AP$_{50}$ (\%)}} && \multicolumn{2}{c}{\textbf{AP$_{75}$ (\%)}}\\
    	  \cmidrule{3-4}\cmidrule{6-7}\cmidrule{9-10}
     	  & & \ls & \ds && \ls & \ds && \ls & \ds \\
    \hline
     \textbf{Faster R-CNN} & \textbf{R-50-C4}         & 54.6 & \textbf{55.9} && 89.4 & \textbf{91.6} && 60.2 & \textbf{61.1}                \\
     \textbf{Faster R-CNN} & \textbf{R-50-FPN}        & 53.9 & \textbf{54.8} && 90.0 & \textbf{92.5} && \textbf{58.5} & 56.9                \\
     \textbf{Faster R-CNN} & \textbf{R-101-FPN}       & 56.7 & \textbf{58.4} && 90.9 & \textbf{93.5} && 62.7 & \textbf{64.0}                \\
     \textbf{Faster R-CNN} & \textbf{X-101-32x8d-FPN} & 57.5 & \textbf{59.1} && 91.5 & \textbf{94.4} && \textbf{63.6} & 63.1                \\
     \textbf{Mask R-CNN}   & \textbf{R-50-C4}         & 56.3 & \textbf{57.5} && 90.4 & \textbf{92.5} && \textbf{63.0} & 62.2                \\
     \textbf{Mask R-CNN}   & \textbf{R-50-FPN}        & 55.8 & \textbf{56.8} && 91.0 & \textbf{93.4} && \textbf{61.1} & 60.7                \\
     \textbf{Mask R-CNN}   & \textbf{R-101-FPN}       & 57.0 & \textbf{58.9} && 91.1 & \textbf{94.0} && 62.4 & \textbf{64.5}                \\
     \textbf{Mask R-CNN}   & \textbf{X-101-32x8d-FPN} & 57.8 & \textbf{59.2} && 91.6 & \textbf{94.0} && 64.6 & \textbf{64.9}                \\
     \hline
     \multicolumn{2}{c}{\textbf{Average}} & 56.2           & \textbf{57.6}           && 90.7           & \textbf{93.2}           && 62.0          &           \textbf{62.2} \\
    \hline
    \end{tabular}
    }
    \end{center}
    \caption{Average precision on BDD100K \emph{train} set.
    }
    \label{tab:bddtrain_mscoco_weights}
\end{table}

\begin{table*}
    \begin{center}
    
    \begin{tabular}{l l  cc c cc c cc}
    	\toprule
    	  \multirow{2}{*}{\textbf{Model}} & \multirow{2}{*}{\textbf{Backbone}}        & \multicolumn{2}{c}{\textbf{AP (\%)}}       && \multicolumn{2}{c}{\textbf{AP$_{50}$ (\%)}} && \multicolumn{2}{c}{\textbf{AP$_{75}$ (\%)}}\\
    	  \cmidrule{3-4}\cmidrule{6-7}\cmidrule{9-10}
     	  & & \ls & \ds && \ls & \ds && \ls & \ds\\
    \hline
     \textbf{Faster R-CNN} & \textbf{R-50-C4}         & \textbf{58.9} & 58.1 && 93.4 & \textbf{93.5} && \textbf{66.6} & 64.7                \\
     \textbf{Faster R-CNN} & \textbf{R-50-FPN}        & \textbf{58.0} & 56.8 && 93.7 & \textbf{94.2} && \textbf{65.1} & 60.1                \\
     \textbf{Faster R-CNN} & \textbf{R-101-FPN}       & \textbf{61.3} & 60.2 && 94.8 & \textbf{95.1} && \textbf{69.4} & 66.1                \\
     \textbf{Faster R-CNN} & \textbf{X-101-32x8d-FPN} & \textbf{62.0} & 61.2 && 95.4 & \textbf{96.1} && \textbf{70.1} & 66.0                \\
     \textbf{Mask R-CNN}   & \textbf{R-50-C4}         & \textbf{60.6} & 59.5 && 94.0 & \textbf{94.6} && \textbf{69.8} & 65.3                \\
     \textbf{Mask R-CNN}   & \textbf{R-50-FPN}        & \textbf{60.1} & 58.8 && 94.6 & \textbf{95.1} && \textbf{67.9} & 63.4                \\
     \textbf{Mask R-CNN}   & \textbf{R-101-FPN}       & \textbf{61.5} & 60.6 && 95.5 & \textbf{95.9} && \textbf{69.0} & 66.9                \\
     \textbf{Mask R-CNN}   & \textbf{X-101-32x8d-FPN} & \textbf{62.4} & 61.5 && 95.6 & \textbf{96.3} && \textbf{71.2} & 68.3                \\
     \hline
     \multicolumn{2}{c}{\textbf{Average}} & \textbf{60.6} & 59.6 && \textbf{94.6} & 95.1 && \textbf{68.6} & 65.1 \\
    \hline
    \end{tabular}
    \end{center}
    \caption{Average precision on BDD100K \emph{train} set with occluded individuals removed for models trained using MS COCO. There are 1855 and 570 individuals labeled as \ls and \ds respectively.}
    \label{tab:bddtrain_mscoco_no_occluded}
    
\end{table*}

\input{loss.tex}

%% file: loss.tex
\subsection{Loss Function for Faster R-CNN and our Reweighting}

Faster R-CNN utilizes a network which suggests regions where an an object is likely to be, known as a Region Proposal Network (RPN). These proposed regions serve as an input to a separate detection network, which is then used for both classification and bounding box regression.

Faster R-CNN has two different objectives, which correspond to the RPN and the separate objection detection network; however, we will solely focus on the latter. Faster R-CNN utilizes a set of predefined boxes of different shapes and sizes, defined as \textit{anchors}, to serve as priors for detecting objects. Within each proposed region, each anchor produces both a probability distribution over the set of candidate classes and a set of regression offsets for the bounding boxes.

The full object detection loss can be written as:
\begin{eqnarray}
  L(\{p_i\}, \{t_i\}) &= \frac{1}{N_{\text{cls}}} \sum_i L_{\text{cls}} (p_i, p_i^*) \\
    &+ \frac{\lambda}{N_{\text{cls}}} \sum_i p_{i}^* L_{\text{reg}}(t_i, t_i^*) \nonumber
 \end{eqnarray}

where $p_i$ is the predicted probability distribution corresponding to a positive anchor, and $p_i^{*}$ corresponds to the one-hot vector corresponding to the ground truth class. $t_i$ and $t_i^{*}$ represent the set of values that parameterize the predicted bounding box and the ground truth bounding box with respect to the positive anchor as such:

\begin{equation*}
    \begin{aligned}[c]
        t_i^x &= \frac{x - x_a}{w_a} \\
        t_i^y &= \frac{y - y_a}{h_a} \\
        t_i^w &= \log \Big ( \frac{w}{w_a} \Big ) \\
        t_i^h &= \log \Big ( \frac{h}{h_a} \Big )
    \end{aligned}
    \begin{aligned}[c]
        t_i^{*x} &= \frac{x^* - x_a}{w_a} \\
        t_i^{*y} &= \frac{y^* - y_a}{h_a} \\
        t_i^{*w} &= \log \Big ( \frac{w^*}{w_a} \Big ) \\
        t_i^{*h} &= \log \Big ( \frac{h^*}{h_a} \Big )
    \end{aligned}
\end{equation*}

\{$x$, $y$\},  \{$x^*$, $y^*$\}, \{$x_a$, $y_a$\} correspond to the centers of the predicted, ground truth, and anchor boxes, and \{$w$, $h$\},  \{$w^*$, $h^*$\}, \{$w_a$, $h_a$\} correspond to the width and height of these boxes.

$L_{\text{cls}}$ is the Log Loss between the set of $k$ classes specified, defined as:

\begin{equation}
    L_{\text{cls}}(p_i, p_i^*) = - \sum_k p_i^{*} \log p_i
\end{equation} 

and $L_{\text{reg}}$ is the smooth $L_1$ loss between the parameterization of the predicted and ground truth boxes. This is written as:

\begin{equation} 
L_{\text{reg}}(t_i, t_i^*) = \sum_{c \in \{x, y, w, h\}} L_{1_\text{smooth}} (t_i - t_i^{*})
\end{equation}
where
\[
L_{1_\text{smooth}}(x) = \begin{cases} 
      0.5 x^2 & \text{if } |x| < 1 \\
      |x| - 0.5 & \text{otherwise}
   \end{cases}
\]

Both $N_{\text{cls}}$ and $N_{\text{reg}}$ are normalization parameters, corresponding the the size of sampled anchors and the number of anchor locations respectively. $\lambda$ is a balancing parameter for the objective. We use the default parameters in the implementation we used for training.

\paragraph{Augmented Loss}
Given a set of attributes \{\ls, \ds, Not a person, A person (cannot determine skin color)\} for members of the $person$ class, we would like to weight the objective based off attribute membership. We can reparameterize our functions, introducing a weight vector, $\mathcal{W} \in \mathbb{R}^4$:
\begin{eqnarray}
  L(\{p_i\}, \{t_i\}, \{a_i\}) =& -\frac{1}{N_{\text{cls}}} \sum_i L_{cls} (p_i, p_i^{*}, \mathcal{W}_{a_i})  \\ & +\frac{\lambda}{N_{\text{reg}}} \sum_i p_{i}^* L_{reg}(t_i, t_i^*, \mathcal{W}_{a_i}) \nonumber
\end{eqnarray} 
where $a_i$ represents the index of the corresponding attribute for a given instance. Therefore, $L_{\text{cls}}$ would become:
 \[ L_{\text{cls}}(p_i, p_i^{*}, \mathcal{W}_{a_i}) = - \mathcal{W}_{a_i} \sum_k p_i^{*} \log ( p_i )\]
 and,
 \[ L_{\text{reg}}(t_i, t_i^*, \mathcal{W}_{a_i}) =  \sum_{c \in \{x, y, w, h\}} \mathcal{W}_{a_i} L_{1_\text{smooth}} (t_i^c - t_i^{*^c})\]

\subsection{Implementation Details}

We used a PyTorch implementation of Faster R-CNN \citep{massa2018mrcnn} for all of the experiments listed. All training was done on 1 NVIDIA V100. All default parameters from the PyTorch implementation were used, except the learning rate, batch size, and the step learning schedule were modified to suit the new dataset. We used a learning rate of 0.01 with a mini batch size of 8 images. Additionally, we used a step learning schedule that decays at 0.1 at iterations 2,233 and 2,792. Each of the BDD100K experiments were run for a total of 3,350 iterations.